\pdfoutput=1

\documentclass[11pt]{article}

\usepackage[]{acl}

\usepackage{times}
\usepackage{latexsym}

\usepackage[T1]{fontenc}

\usepackage[utf8]{inputenc}

\usepackage{microtype}

\usepackage{inconsolata}

\usepackage{graphicx}
\usepackage{xcolor}
\usepackage{booktabs}
\usepackage{amsmath}
\usepackage{subfig}
\usepackage{enumitem}
\usepackage{bbm}
\usepackage{titlesec}

\titlespacing*\section{0pt}{7pt plus 2pt minus 2pt}{1pt plus 2pt minus 1pt}
\titlespacing*\subsection{0pt}{7pt plus 2pt minus 2pt}{1pt plus 2pt minus 2pt}
\titlespacing*\subsubsection{0pt}{7pt plus 2pt minus 2pt}{1pt plus 2pt minus 2pt}

\newcommand\datalink{https://github.com/google/wmt-mqm-human-evaluation/tree/main/generalMT2022}

\newcommand\favmetric{Stable Ranking Probability}
\newcommand\favmetricabbr{SRP}
\newcommand\rqgroup{How should rateable items be grouped?}
\newcommand\rqload{How should workload be distributed?}
\newcommand\rqnorm{How should scores be normalized?}
\newcommand\rqtotal{How many items should be annotated?}
\newcommand\rqrpi{How many ratings per item are needed?}
\setlist[itemize]{noitemsep,topsep=4pt}

\title{Finding Replicable Human Evaluations via Stable Ranking Probability}

\author{Parker Riley~~~{\bf Daniel Deutsch}~~~{\bf George Foster} \\ {\bf Viresh Ratnakar}~~~{\bf Ali Dabirmoghaddam}~~~{\bf Markus Freitag} \\ Google}

\begin{document}
\maketitle
\begin{abstract}
Reliable human evaluation is critical to the development of successful natural language generation models, but achieving it is notoriously difficult. Stability is a crucial requirement when ranking systems by quality: consistent ranking of systems across repeated evaluations is not just desirable, but essential. Without it, there is no reliable foundation for hill-climbing or product launch decisions.
In this paper, we use machine translation and its state-of-the-art human evaluation framework, MQM, as a case study
to understand how to set up
reliable human evaluations that yield stable conclusions. We investigate the optimal configurations for item allocation to raters, number of ratings per item, and score normalization. Our study on two language pairs provides concrete recommendations for designing replicable human evaluation studies.
We also collect and release the largest publicly available dataset of multi-segment translations rated by multiple professional translators, consisting of nearly 140,000 segment annotations across two language pairs.\footnote{\datalink}
\end{abstract}

\section{Introduction}
When conducting an evaluation to rank natural language generation (NLG) systems, including modern ``generative AI'' systems, the goal is to reliably recover the ``true'' ranking of the systems by quality on the tasks/domains of interest. To date, human evaluation remains the best way to do this. However, human raters can exhibit different behaviors when rating NLG outputs. For example, raters can have different stylistic preferences, and some may grade more leniently or harshly than others. Rigorous rater training procedures and precise annotation guidelines are helpful in addressing this but are not guaranteed to completely eliminate rater differences. Coupled with the fact that any given study will evaluate only a subset of potential NLG outputs of interest using a finite set of raters, decisions around which raters rate which items can alter the final system ranking. This inherent instability can make it difficult to confidently attribute ranking changes between evaluations to changes in the underlying systems.

Our goal in this work is to offer recommendations on various aspects of designing multi-system NLG human evaluation studies. We propose evaluating results through the lens of \textbf{stability}: the degree to which a specific evaluation methodology produces the same system ranking when repeated. Specifically, we seek to answer these questions:
\begin{itemize}
    \item \rqgroup
    \item \rqload
    \item \rqrpi
    \item \rqnorm
    \item \rqtotal    
\end{itemize}

To do this, we use machine translation (MT) evaluation as a case study. For MT, Multidimensional Quality Metrics (MQM) is the state-of-the-art human evaluation framework \citep{lommel-etal-mqm,freitag-etal-2021-experts}. In MQM, expert raters identify error spans within translations, which are automatically converted to numeric scores. Section~\ref{sec:mqm} contains additional details.

Using MT and MQM as a specific use-case, we make the following contributions:

\begin{itemize}
    \item We provide concrete recommendations for designing NLG system ranking evaluations (\S\ref{sec:recommendations}).
    \item We propose a meta-evaluation metric (\S\ref{sec:replicability}) and framework (\S\ref{sec:sim_analysis}), based on the notion of stability, for evaluating evaluation methodologies.
    \item We justify our recommendations by analyzing two MQM datasets that we collected (\S\ref{sec:results}).
    \item We publicly release nearly 140,000 segment ratings, comprising the largest publicly available dataset of multi-segment translations rated by multiple professional translators.
\end{itemize}
\section{Terminology}\label{sec:terminology}
A few key terms used throughout this work are defined as follows:
\begin{itemize}
    \item \textbf{Segment}: A unit of one or occasionally multiple sentences, either in the input or the output.
    \item \textbf{Document}: A sequence of input segments (e.g. an excerpt from a news article).
    \item \textbf{Item}: One MT system's output on a single document.
    \item \textbf{Bucket}: A subset of items in one of our datasets that were all evaluated by the same set of raters.
\end{itemize}
\section{Research Questions and Findings}\label{sec:recommendations}
Here we pose the research questions considered in this work. Each is expressed in a way that seeks a recommendation on how to implement some facet of an NLG ranking study. We provide these 
recommendations below, and justify them experimentally in Section~\ref{sec:results} based on their positive effect on stability, after describing our
meta-evaluation methodology in the intervening sections.

\textbf{\rqgroup} Some types of NLG evaluation use a side-by-side methodology, where raters are shown an input and two outputs from different systems, and then assess their relative quality. In contrast, raters only see one output at a time in the MQM framework that we use for our analysis. Thus, the closest analogue is what we term a \emph{pseudo-side-by-side} (pSxS) methodology, which means that all system outputs from a given input are assigned to the same rater. 

Based on our results, we strongly recommend use of the pSxS methodology.

\textbf{\rqload} When sending a collection of rating items to a pool of raters, one consideration is whether to limit how many items any single rater can annotate. Doing so can slow down progress as fast raters hit their limits, but in exchange for preventing results from being skewed toward any particular rater.

We recommend that all raters be given an equal share of the total rating workload. However, we observe that excessive rating noise from differences in both rater behaviors and input documents can affect this recommendation, so we couple it with a recommendation to leverage clear annotation guidelines and rater training to limit noise of this type. Section~\ref{sec:balancing_results} further discusses the effect of this noise.

\textbf{\rqnorm} One potential technique for controlling noise from differences in rater behavior is to apply normalization to collected ratings. This is especially attractive as an option to address noise introduced in the setup of an evaluation after it has already been run.

We weakly recommend applying rater-wise Z-score normalization. We make this recommendation weakly because the most pronounced gains of Z-score normalization occur in settings that run counter to
other recommendations provided here.

\textbf{\rqtotal} For NLG it is generally infeasible to evaluate system performance on all possible inputs of interest, and practitioners must select a comparatively small subset to be evaluated. When allocating budgets, it is useful to understand how much reliability is gained by annotating additional items.

We do not make a concrete recommendation here, in part because judging the trade-off between reliability and cost is highly setting-dependent. However, we show all of our results over a wide range of item counts to assist practitioners in making this judgment for themselves.

\textbf{\rqrpi} Collecting more ratings for each item can limit noise from differences in rater behavior, but in a practical context with fixed budgets, this generally requires reducing the number of distinct items proportionally, introducing additional noise from item-level differences. The conclusion therefore depends on which source of noise is greater.

We recommend that each item be annotated by a single rater to allow for the maximum number of distinct items given a fixed budget. Note that evaluation settings other than MQM for MT may have different noise profiles that affect this recommendation (see Section~\ref{sec:limitations}).
\section{Data}\label{sec:data}
\subsection{Background on MQM}\label{sec:mqm}

In the Multidimensional Quality Metrics (MQM) framework for human evaluation of machine translation (MT), expert annotators identify error spans within translations and assign a hierarchical category (e.g. Fluency/Grammar or Accuracy/Mistranslation) and severity (Major or Minor).
Unlike the Likert-style scheme used in many previous studies, e.g. \citet{graham-etal-2020-assessing}, MQM raters do not directly assign scalar scores; instead, scores are calculated after the annotation step by applying a weighting scheme that considers severity and category, and systems are ranked by their average score over segments in the evaluation set. The framework does not use reference translations, and each rating item consists of the input segments of one document and the corresponding translation from a single system.

\subsection{Dataset Descriptions}
For our analysis, we use a set of MQM annotations that we collected for English-German and English-Chinese MT data from the WMT22 evaluation campaign \citep{kocmi-etal-2022-findings} and metric shared task \citep{freitag-etal-2022-results}. Each language pair's data comprises 181 documents, each of which is broken down into segments which are individually scored by raters (in the context of the documents in which they occur). Each segment is generally one sentence, but sometimes several. The same 181 source documents were used for both datasets, but for English-German they were truncated to the first 10 segments.\footnote{This is inherited from the WMT22 metric shared task.} For every input document, all system outputs in a given language were evaluated by the same 3 raters, from a pool of 7 raters for English-German and 6 for English-Chinese. All raters are professional translators that regularly perform MQM annotation and are paid fair market wages. Basic statistics about the datasets are shown in Table~\ref{tab:basic_stats}. Combined, our two datasets include nearly 140,000 segment ratings.

Both datasets are divided into \textbf{buckets}, which are sets of documents rated by the same 3 raters. No two buckets share the exact same set of 3 raters. We experiment with a different bucketing scheme for each dataset. For English-German, there are 7 buckets of 25-26 documents, and each rater rated 3 buckets. For English-Chinese there are only 2 buckets of 90-91 documents, and each rater rated a single bucket. Appendix~\ref{app:buckets} has further details.

Mean MQM scores for all systems are listed in Appendix~\ref{app:mean_scores}.

\begin{table}[htb]
\centering
\footnotesize
\begin{tabular}{lrr}
\toprule
&\textbf{English-} & \textbf{English-}\\
&\textbf{German} & \textbf{Chinese}\\
\midrule
\# Documents & 181 & 181\\
\# Segments & 1315 & 2037\\
Min \# Segments/Doc & 1 & 1\\
Max \# Segments/Doc & 10 & 51\\
\# Unique Raters & 7 & 6\\
\# Systems & 15 & 13\\
\bottomrule
\end{tabular}
\caption{Basic counts for each dataset.}\label{tab:basic_stats}
\end{table}

\subsection{Rater Behavior and Agreement}\label{sec:behavior}

\begin{figure}[htb]
    \centering
        \centering
        \includegraphics[width=0.45\textwidth]{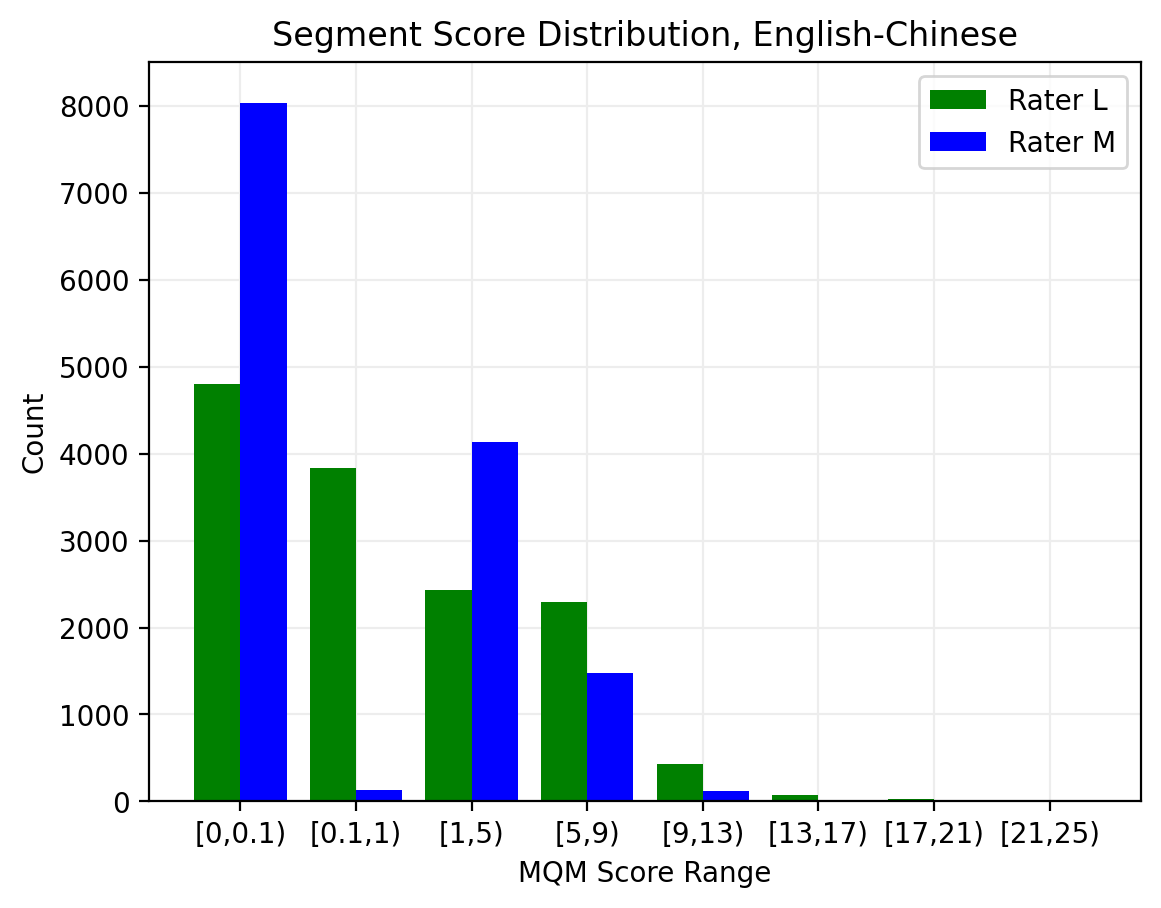}
        \caption{Distribution of segment-level MQM scores for two English-Chinese raters who rated the same items, highlighting differences in rater behavior.}
        \label{fig:rater_behavior}
\end{figure}

\begin{table}[htb]
    \centering
    \footnotesize
    \begin{tabular}{lrr}
    \toprule
          & \textbf{English-} & \textbf{English-} \\
         \textbf{Aggregation} & \textbf{German} & \textbf{Chinese} \\
         \midrule
         Single Documents & 0.40 & 0.29 \\
         All Shared Documents & 0.69 & 0.85 \\
         \bottomrule
    \end{tabular}
    \caption{Average over rater pairs of either: the average Kendall's Tau between document-level system rankings or the Kendall's Tau between system rankings over all documents rated by both raters.}
    \label{tab:pairwise_corr}
\end{table}

As previously mentioned, human raters can exhibit different behaviors when evaluating NLG tasks, for reasons such as preferences and leniency. These effects are present in our datasets, and we briefly explore them here.

We find that raters often have different distributions of segment-level MQM scores (calculated from the severity/category information of error spans they annotated). Figure~\ref{fig:rater_behavior} illustrates this for two raters who rated the same items. Based on manual inspection of the data, we hypothesize that these differences are not generally due to a rater performing the task incorrectly, but rather due to differences in harshness or leniency between raters: a Minor error to one rater may be a Major error to another.

Table~\ref{tab:pairwise_corr} summarizes the average agreement between rater pairs, calculated either as the average Kendall's Tau between single-document rankings or the Kendall's Tau between rankings calculated from all documents that both raters annotated.
We point out that these two levels of granularity disagree on which dataset has higher agreement. This indicates that while the English-Chinese raters tended to agree on the overall quality ranking of systems more than the English-German raters, there was more variance in quality rankings of individual documents.

\section{Quantifying Replicability}\label{sec:replicability}
In human evaluation research, it is often difficult to define a ``ground truth'' to compare to, as we are in some sense attempting to determine what to use as the ground truth in the first place. However, an important feature of an evaluation methodology is replicability, which in the context of system ranking studies we examine through the lens of \textbf{stability}. We imprecisely define stability as the tendency of an evaluation methodology to produce the same system rankings when repeated. Stability is a critical property of a sound evaluation methodology, because unstable methodologies make it difficult to attribute changes in system rankings to real changes in quality.

We propose to estimate the stability of a methodology using a metric calculated on pairs of studies designed with that methodology. We experimented with approximately a dozen different metrics to quantify stability, including well-known correlation measures like average Pearson correlation and average Kendall's Tau. All of them pointed toward the same answers to our research questions, so we present only one metric in the main body of this work: \textbf{\favmetric~(\favmetricabbr)}.

Given a set of studies with a common methodology, the \favmetric~of that set is the proportion of ordered study pairs for which \textit{all} significant system pair differences in the first study are reflected in the rankings of the second study, ignoring significance. That is, 
if a study finds that system A is significantly better than system B and that system C is significantly better than system D, \favmetricabbr~measures the probability that repeating the study with the same methodology will agree that A is better than B \textit{and} that C is better than D. This can be expressed with the following equations:
\begin{multline*}
    \mathrm{SRP}(E) = \frac{1}{|\pi_2(E)|} \sum_{(e_1,e_2) \in \pi_2(E)} \mathrm{SR}(e_1,e_2)\\
    \mathrm{SR}(e_1,e_2) = \mathbbm{1}(\forall i,j~\mathrm{SB}_{e_1}(s_i,s_j) \Rightarrow \mathrm{B}_{e_2}(s_i,s_j))
\end{multline*}
where $E$ is a set of studies, $\pi_2(E)$ is the set of all ordered pairs (i.e. size-2 permutations) of studies in $E$, $\mathbbm{1}(p)$ is $1$ if predicate $p$ is true and $0$ otherwise, $\mathrm{SB}_{e_1}(s_i, s_j)$ is a boolean function representing whether system $s_i$ is statistically significantly better than system $s_j$ according to study $e_1$, and $\mathrm{B}_{e_2}(s_i, s_j)$ represents whether $s_i$ is better than $s_j$ according to study $e_2$ without considering significance.

\favmetricabbr~is fairly strict in that it requires directional agreement over all significant system pair differences in a given study, with no partial credit for agreement on only some system pairs\footnote{We explored a version that requires significance in both studies, but found it to be \textit{too} strict, consistently near 0.}. We find this to be important because we observe that many system pairs in our datasets have such large quality differences that almost all simulated studies detect a significant difference, and we want to focus on the more difficult case of reliably detecting differences between systems with similar quality. We found that other metrics such as average Kendall's Tau tend to be dominated by these easily-distinguishable system pairs, yielding a narrow range of values. Another benefit of \favmetricabbr~is that it is interpretable because it is a probability.

To determine statistical significance between mean MQM scores for two systems, we use a random permutation test on segment scores, with the caveat that system labels for segments in the same document are always permuted as a group. We always use 500 permutations, and conclude that a difference is significant if the $p$-value is less than or equal to $\alpha=0.05$.

\section{Simulation Analysis}\label{sec:sim_analysis} 
Our approach to answering the questions listed in Section~\ref{sec:recommendations} is to simulate a collection of studies sharing a common evaluation methodology by repeatedly sampling a subset of ratings from our datasets. We then evaluate the replicability of that methodology using our \favmetricabbr\ stability metric.

\subsection{Evaluation Methodology Features}\label{sec:features}
We formalize aspects of evaluation methodology as features which take specific values. We organize the description of each feature based on the research question that it helps us answer.

\subsubsection{\rqgroup}

We explore this question using a feature we call \textbf{Item Grouping}, which describes constraints on which items must be rated by the same raters. The feature takes one of three values:
\begin{itemize}
    \item \textit{Pseudo-Side-by-Side (pSxS)}: All system outputs from a given input document form a group where all are annotated by the same rater(s).
    \item \textit{System-Balanced}: Groups of one output from each system are formed and annotated by the same rater(s) (as above), but the outputs can be from different input documents.\footnote{A similar technique was used by \citet{bojar-etal-2018-findings}, but without comparison to alternatives.}
    \item \textit{No Grouping}: No constraints are placed (i.e. raters may be assigned more outputs from some systems than others).
\end{itemize}

Compared to the unconstrained setting, system-balanced item grouping controls for noise arising from input-agnostic differences in rater behavior: with no constraints, most outputs from one system might be assigned to a particularly lenient rater, boosting that system's rating. Using pSxS item grouping further controls for noise arising from input-dependent differences, such as a rater having strict stylistic preferences in a particular domain.

\subsubsection{\rqload}
The feature we use for this question is \textbf{Load Balancing}, which governs how equally the total annotation workload is distributed among the raters. The possible feature values are:
\begin{itemize}
    \item \textit{Fully Balanced}: Items are distributed as evenly as possible among raters.
    \item \textit{Entropy-Balanced}: Parameterized by an entropy target between 0 and 1, the normalized entropy of the workload distribution over raters is within $0.03$ of the specified target.
\end{itemize}

The normalized entropy is defined as:
$-\sum_{r \in \mathcal{R}} p(r) \log p(r) / \log|\mathcal{R}|$
where $R$ is the set of raters in the given dataset, and $p(r)$ is the proportion of total item ratings assigned to rater $r$. Normalized entropy will be $1$ when every rater is assigned the same number of items, and $0$ when a single rater is assigned all items. The fully balanced setting is a special case of the entropy-balanced setting when the entropy is maximal, but not necessarily 1 when items cannot be distributed completely evenly. Note that the difference in bucketing between our two datasets means that the minimal entropy that we can actually instantiate is $0.51$ for English-German and $0.38$ for English-Chinese.

While real studies are unlikely to \textit{explicitly enforce} a moderately imbalanced workload distribution, moderate imbalance is a likely result of studies that place no restrictions on how many items each rater evaluates, which is the default setup for many evaluation platforms (including Amazon Mechanical Turk). Our Load Balancing feature allows us to measure the potential stability impact of this.

\subsubsection{\rqnorm}
We call this feature \textbf{Normalization}, which describes what normalizing transformation is applied to segment-level quality scores after annotation and before determining the overall system ranking. We examine four values of this feature:
\begin{itemize}
    \item \textit{Unnormalized}: No normalization is applied.
    \item \textit{Mean-Normalized}: Every score is multiplied by a rater-specific value that results in all raters having the same mean MQM score, equal to the overall pre-normalized mean MQM score for the entire study.
    \item \textit{Error-Normalized}: Scores are first mean-normalized, then multiplied by a rater-specific value that is proportional to the total \textit{number of errors} identified by that rater, ignoring severity. The value is chosen such that the mean MQM score for the entire study remains unchanged.
    \item \textit{Z-Score-Normalized}: Each rater's mean score is subtracted from each of their ratings, and each result is divided by the standard deviation of the rater's scores.
\end{itemize}

A common motivation for all above normalization methods is that the scores from harsh raters will have greater magnitude than from lenient raters, and will therefore dominate the final system ranking computed from mean scores. If we trust all raters equally then this is undesirable; mean-normalized and Z-score-normalized studies avoid this by making all raters have the same mean score. Z-score-normalized studies additionally control for differences in the range of scores that raters assign.

Error-normalized studies partially reverse the assumption of trusting all raters equally, based on the hypothesis that expert MQM raters have high precision but variable recall over ``true'' errors, so ratings with more identified errors are likely to be of higher quality. Note that this method still ignores differences in rater harshness from use of Major vs. Minor severities.

\subsubsection{\rqtotal}\label{sec:num_docs}
We examine this question with two features that interact with each other. The first, called \textbf{Number of Documents}, simply denotes how many input documents are included in a study. We always include all system outputs from each included document.

When simulating studies with a strict subset of available documents, if we were to use the \textit{same} subset of documents for all studies, then our results could be overly dependent on the specific subset that was selected. We instead resample the document subset with some frequency, described by the \textbf{Document Resampling} feature, which takes one of two values:
\begin{itemize}
    \item \textit{Resampled Documents}: The document subset is resampled for every simulated study.
    \item \textit{50 Simulations / Document Set}: 50 studies are simulated before resampling.
\end{itemize}

This feature interacts with our calculation of \favmetric. With 50 simulations per document set, we restrict the \favmetricabbr~calculation to only consider ordered pairs of studies with the same input documents. This is because practitioners usually iterate on a test/validation set with a fixed set of documents, and calculating agreement between studies on different documents would not accurately reflect this. However, we do not apply this restriction for the one experiment where we use resampled documents (\S\ref{sec:results_numratings}).

\subsubsection{\rqrpi}
We investigate this question with the feature called \textbf{Ratings per Item}, which takes one of two values:
\begin{itemize}
    \item \textit{Single-Rated}: Each item is annotated once.
    \item \textit{Double-Rated}: Each item is annotated twice, by different raters.
\end{itemize}

As mentioned in Section~\ref{sec:recommendations}, we approach this question from the perspective of a practitioner with a fixed annotation budget. Therefore, we control for the \textit{total number of ratings} by comparing single-rated studies of $n$ input documents to double-rated studies of $\frac{n}{2}$ input documents.

\subsection{Simulation Procedure}\label{sec:sim_procedure}
\begin{figure}
    \centering
    \includegraphics[width=0.48\textwidth]{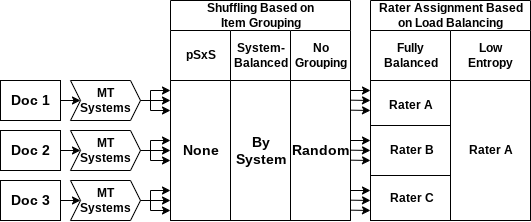}
    \caption{Overview of the simulated rater assignment procedure, focusing on two features. System translations are collected for the shuffled documents, and then optionally shuffled based on the Item Grouping feature. ``By System'' means that items are only shuffled among the positions from the same system. Items are then assigned to shuffled raters, with the (im)balance of the distribution controlled by the Load Balancing feature.}
    \label{fig:sim_procedure}
\end{figure}

When simulating studies by choosing subsets of ratings from our datasets, we first specify the value of each feature described in Section~\ref{sec:features}. These feature values then inform the procedure for simulating studies with those properties. The two most relevant features are Item Grouping and Load Balancing, and their effect on the procedure is illustrated in Figure~\ref{fig:sim_procedure}. Further details are in Appendix~\ref{app:simulation}.
\section{Results}\label{sec:results}

As with Sections~\ref{sec:recommendations} and \ref{sec:features}, we organize this section by research question, with experimental results justifying the recommendations made in Section~\ref{sec:recommendations}. However, for answering ``\rqtotal'' we examine the effect of the Number of Documents feature in all experiments in this section (the $x$-axis in our figures).

\subsection{\rqgroup}
\begin{figure*}[htb]
    \centering
    \subfloat{\includegraphics[scale=0.18]{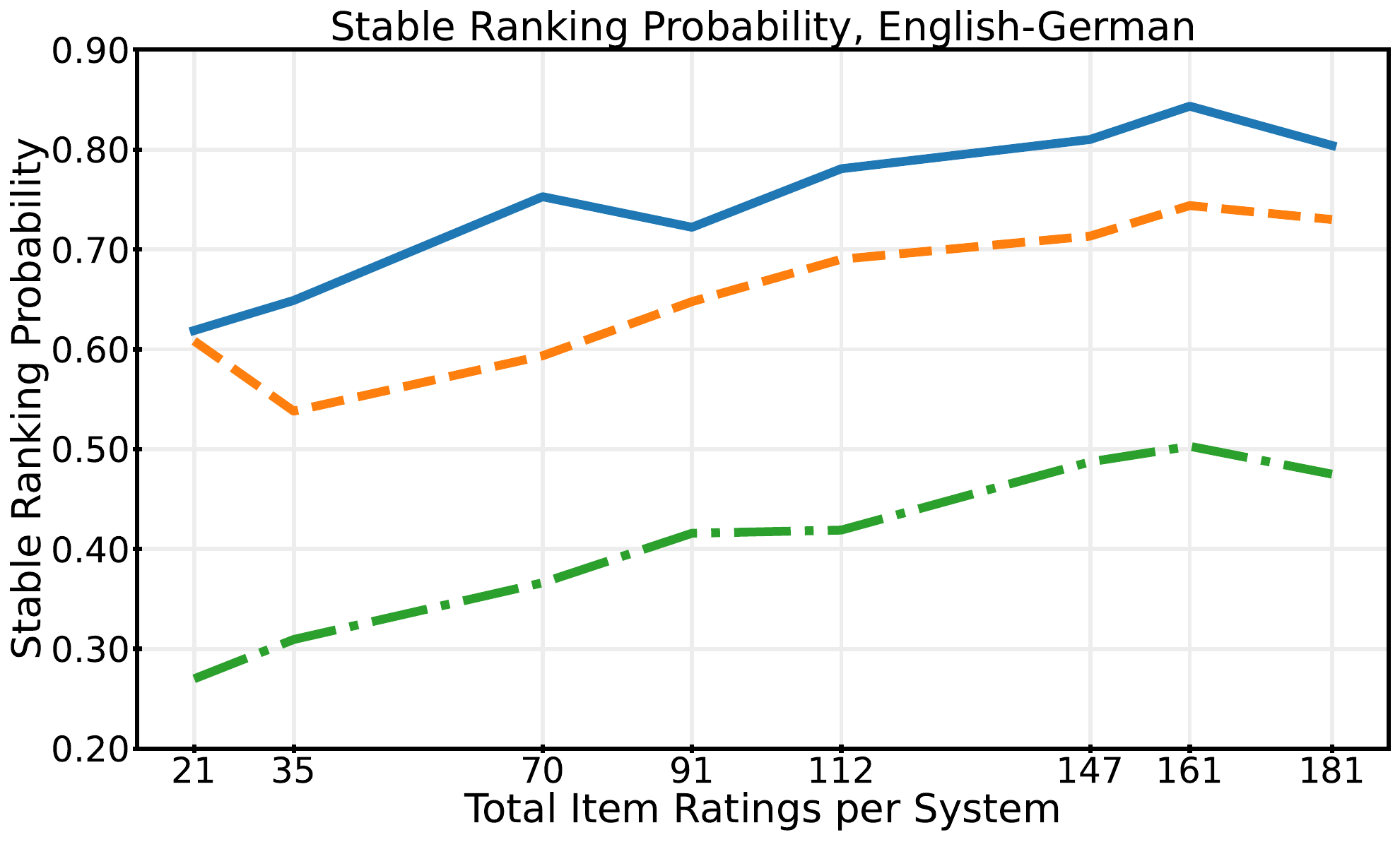}}\hfill
    \subfloat{\includegraphics[scale=0.18]{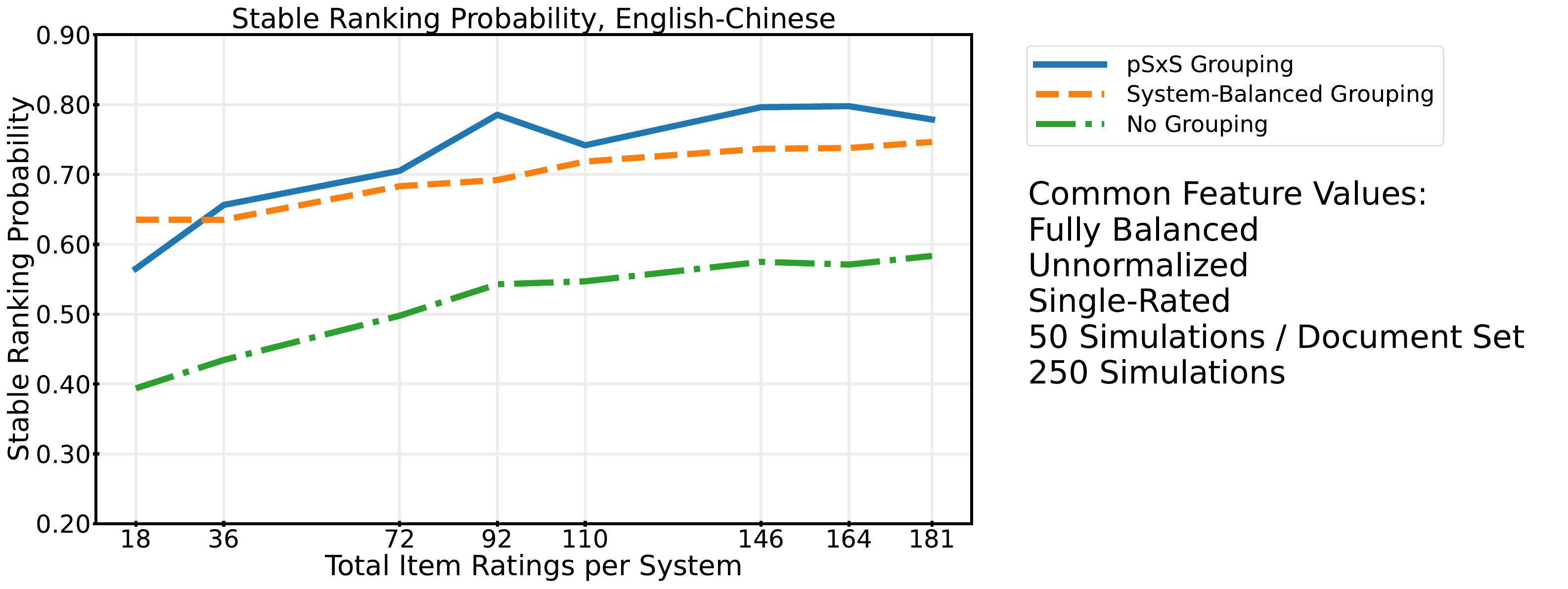}}
    \caption{Evaluation of the Item Grouping feature.}
    \label{fig:psxs}
\end{figure*}

Figure~\ref{fig:psxs} illustrates that grouping items using the pSxS methodology (where all system outputs on the same input are annotated by the same rater) provides a massive stability boost over no grouping at all document counts, approximately doubling stability in the English-German case. Using a system-balanced grouping provides a large improvement to stability over using no grouping, but still generally underperforms the pSxS grouping. Coupled with the simplicity of pSxS grouping, we believe these results strongly support our recommendation.

\subsection{\rqload}\label{sec:balancing_results}
\begin{figure*}[htb]
    \centering
    \subfloat{\includegraphics[scale=0.18]{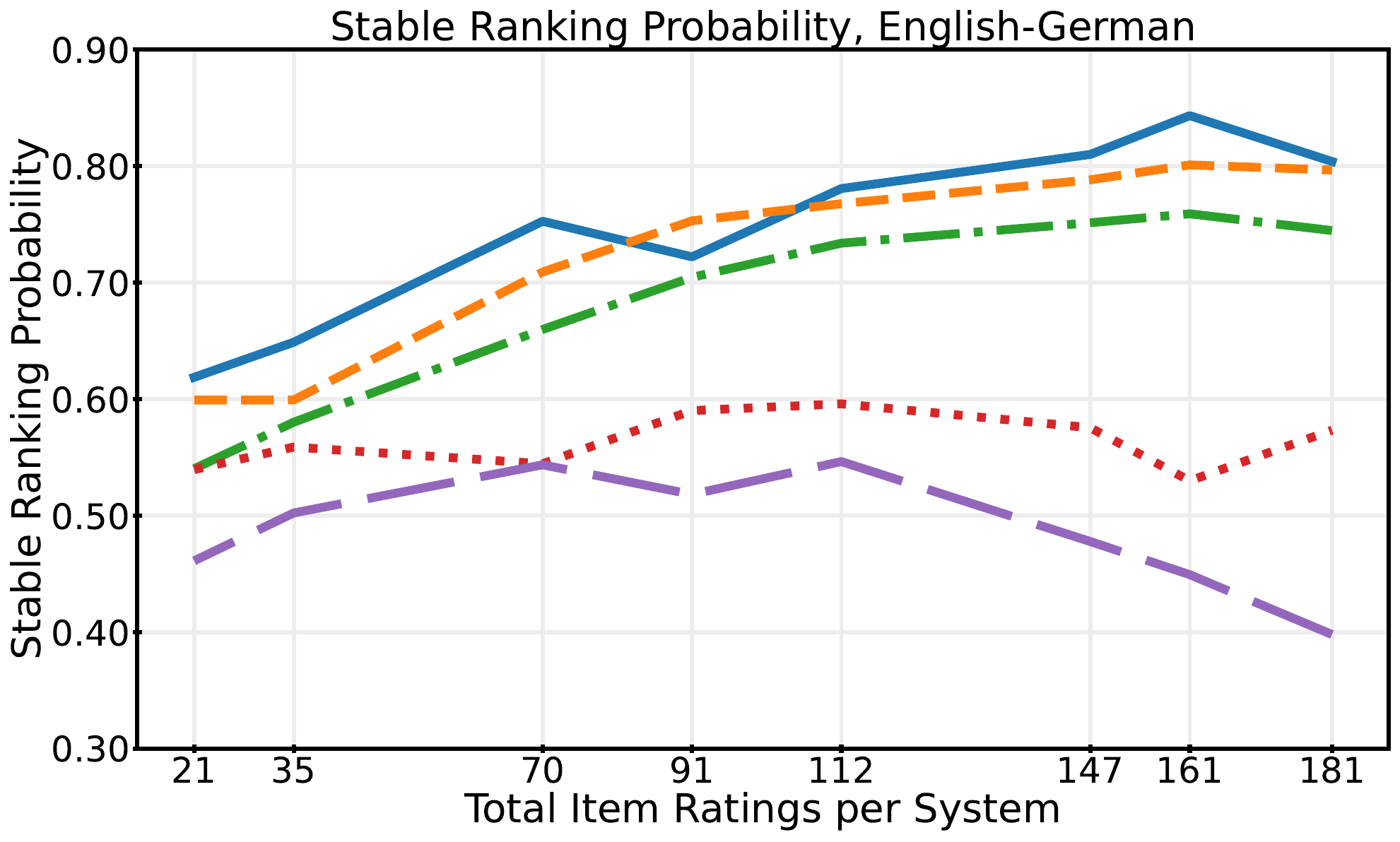}}\hfill
    \subfloat{\includegraphics[scale=0.18]{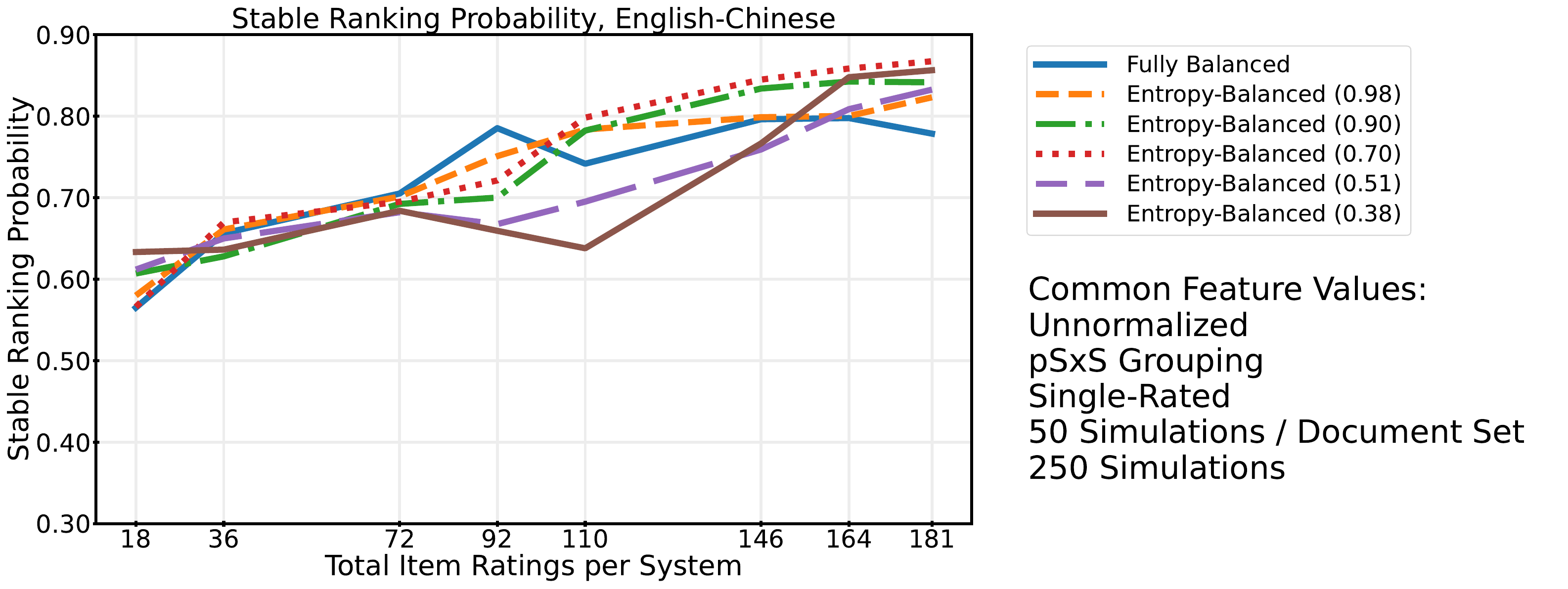}}
    \caption{Evaluation of the Load-Balancing feature. Our datasets show the opposite relationship between imbalance and stability.}
    \label{fig:bal}
\end{figure*}

Figure~\ref{fig:bal} tells two conflicting stories about the relationship between workload imbalance and stability, with English-German results favoring fully balanced studies and English-Chinese results favoring highly imbalanced studies. We attribute this to an interesting quirk of our data, discussed in Section~\ref{sec:behavior}: compared to the English-German raters, pairs of our English-Chinese raters tended to agree more on system rankings calculated over all documents they both rated, despite agreeing \textit{less} on individual documents. This means that low-entropy studies (where each bucket is essentially rated by a single rater) tend to converge to the same result regardless of which raters are selected, while high-entropy studies exhibit more variance because each selected rater sees a smaller portion of the data. This is why we recommend taking steps to limit differences in rater behavior: we believe that low document-level ranking agreement is a sign of excessive noise, and when it is mitigated, fully balanced studies yield higher stability.

\subsection{\rqnorm}
\begin{figure*}[htb]
    \centering
    \subfloat{\includegraphics[scale=0.18]{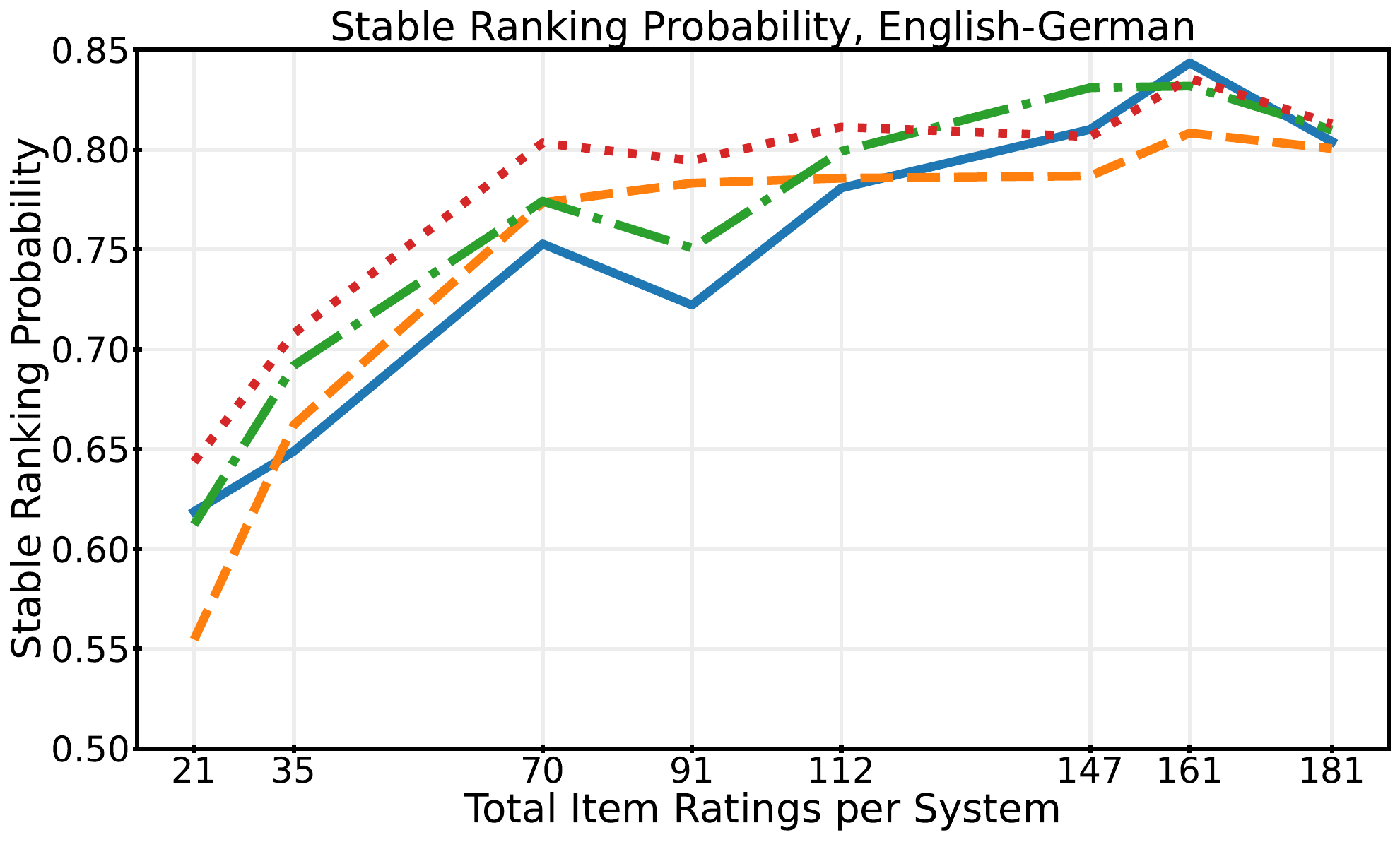}}\hfill
    \subfloat{\includegraphics[scale=0.18]{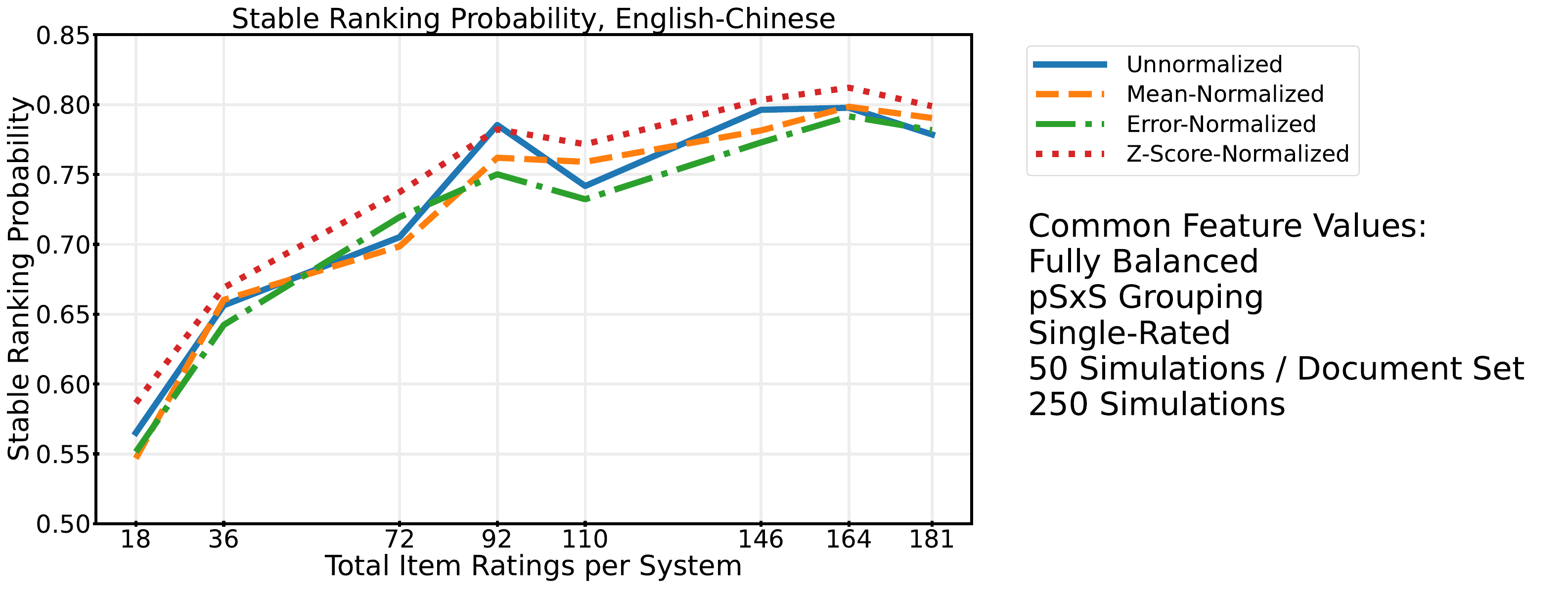}}
    \caption{Evaluation of the Normalization feature in our recommended setting.}
    \label{fig:norm_best}
\end{figure*}

Figure~\ref{fig:norm_best} illustrates the effect of normalization in our recommended setting. We see that Z-score-normalized studies exhibit higher stability than others, but the magnitude of the difference is not always large. In English-German, the gap is quite small at high document counts but larger for smaller studies. In English-Chinese, the gap is generally modest.

\begin{figure*}[htb]
    \centering
    \subfloat{\includegraphics[scale=0.18]{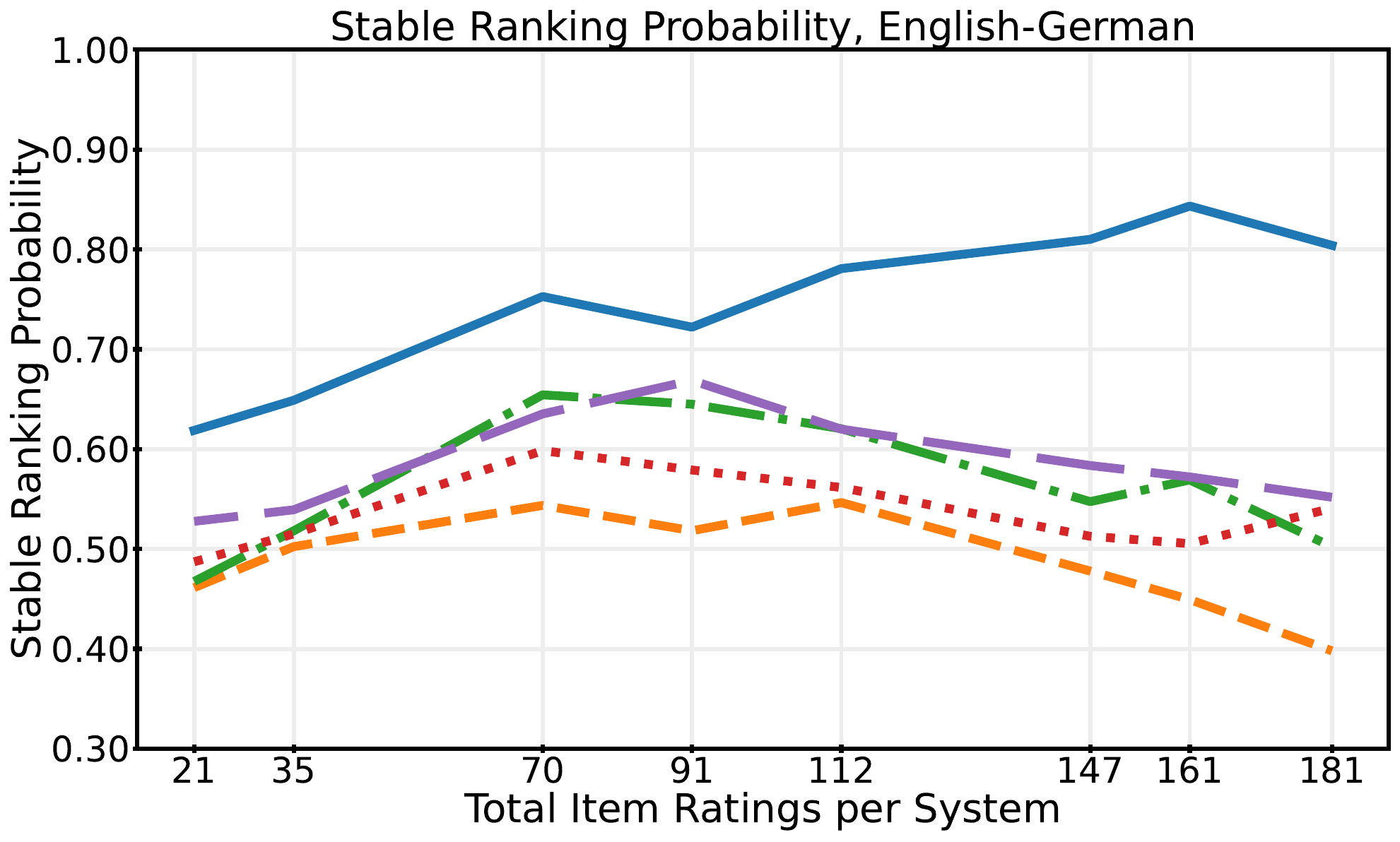}}\hfill
    \subfloat{\includegraphics[scale=0.18]{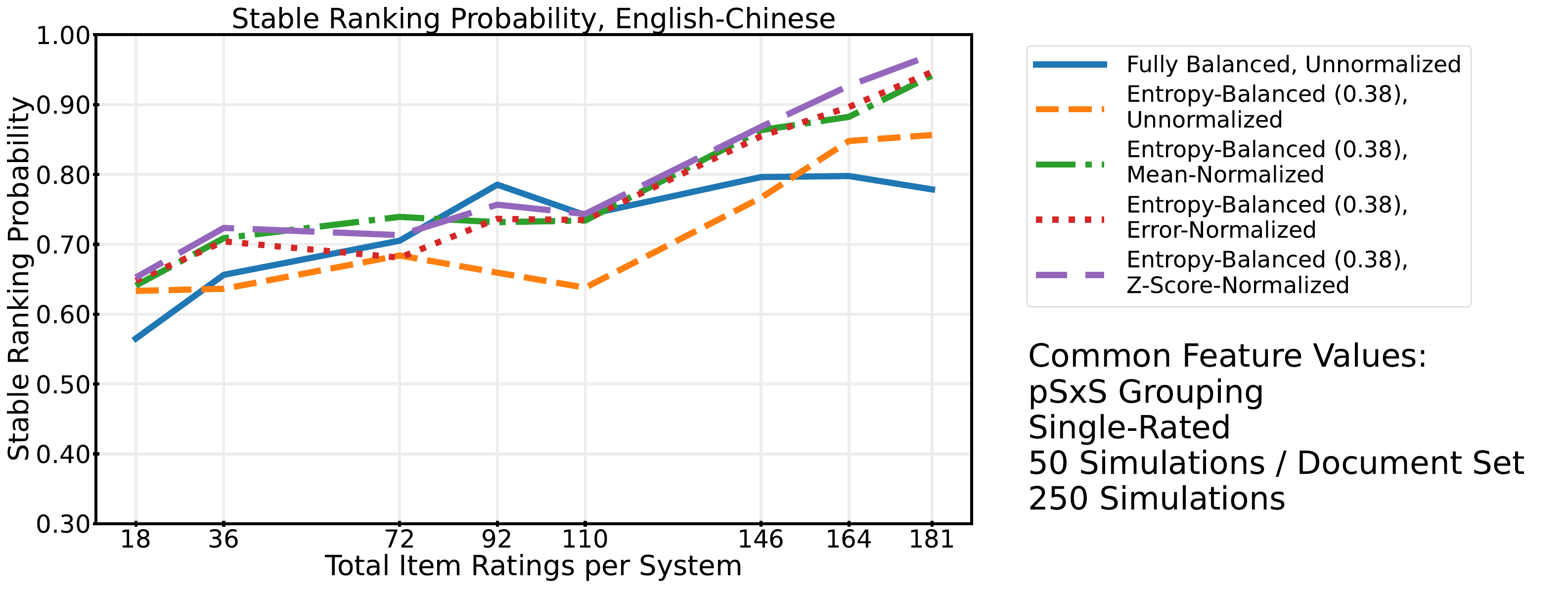}}
    \caption{Evaluation of the Normalization feature in balanced vs. highly imbalanced studies.}
    \label{fig:norm_imbal}
\end{figure*}

Figure~\ref{fig:norm_imbal} examines the effect of normalization on highly-imbalanced studies, with unnormalized fully balanced studies shown for reference. In both cases, Z-Score normalization provides a significant boost to stability. For English-German, this boost does not close the gap below fully balanced studies. For English-Chinese, this boost widens the gap above fully balanced studies. Note that normalizing fully balanced studies does not close the gap to highly-imbalanced studies in English-Chinese (cf. Figures~\ref{fig:norm_best} and \ref{fig:norm_imbal}). The effect of other normalization methods on stability is less consistent.

\begin{figure*}[htb]
    \centering
    \subfloat{\includegraphics[scale=0.18]{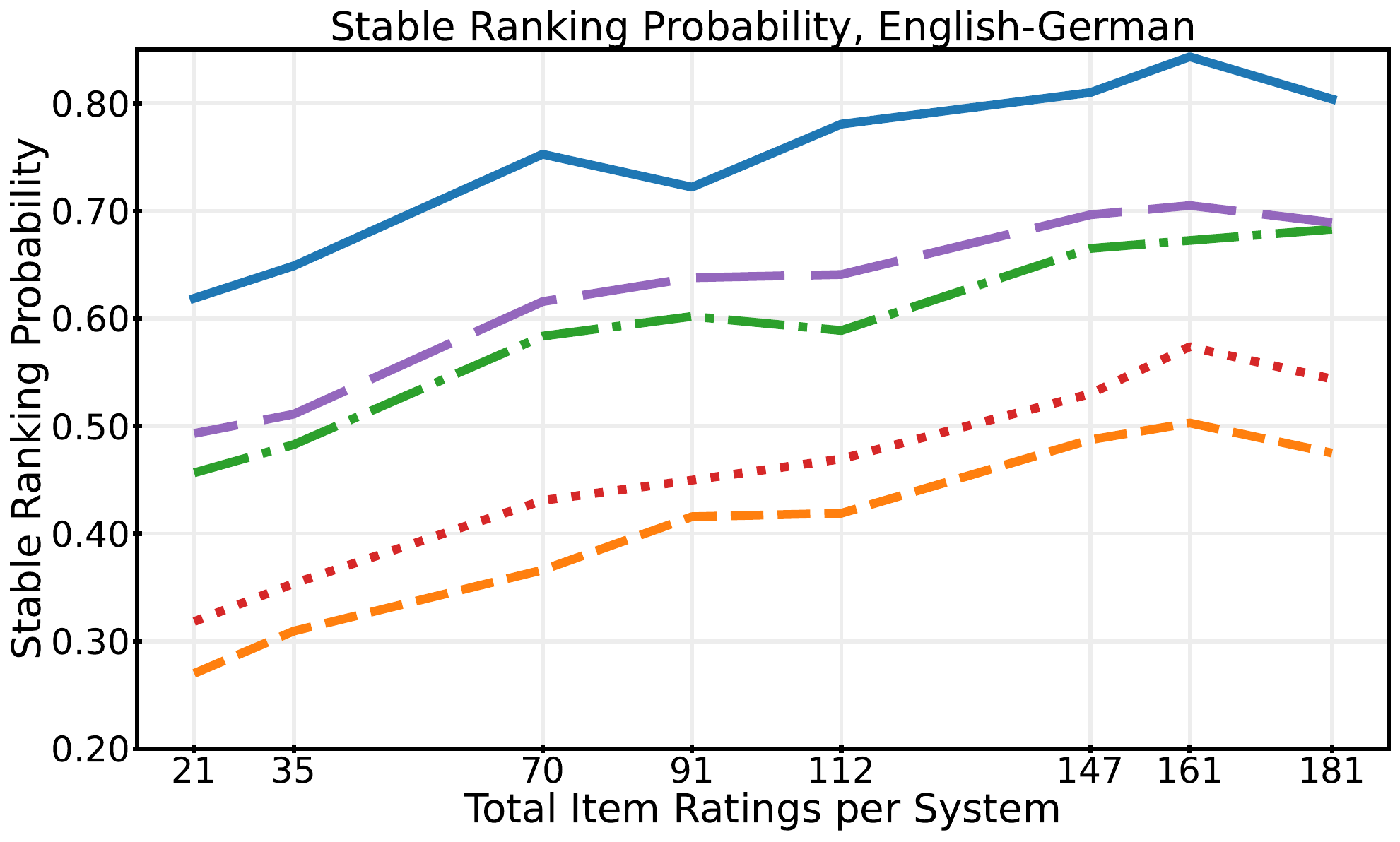}}\hfill
    \subfloat{\includegraphics[scale=0.18]{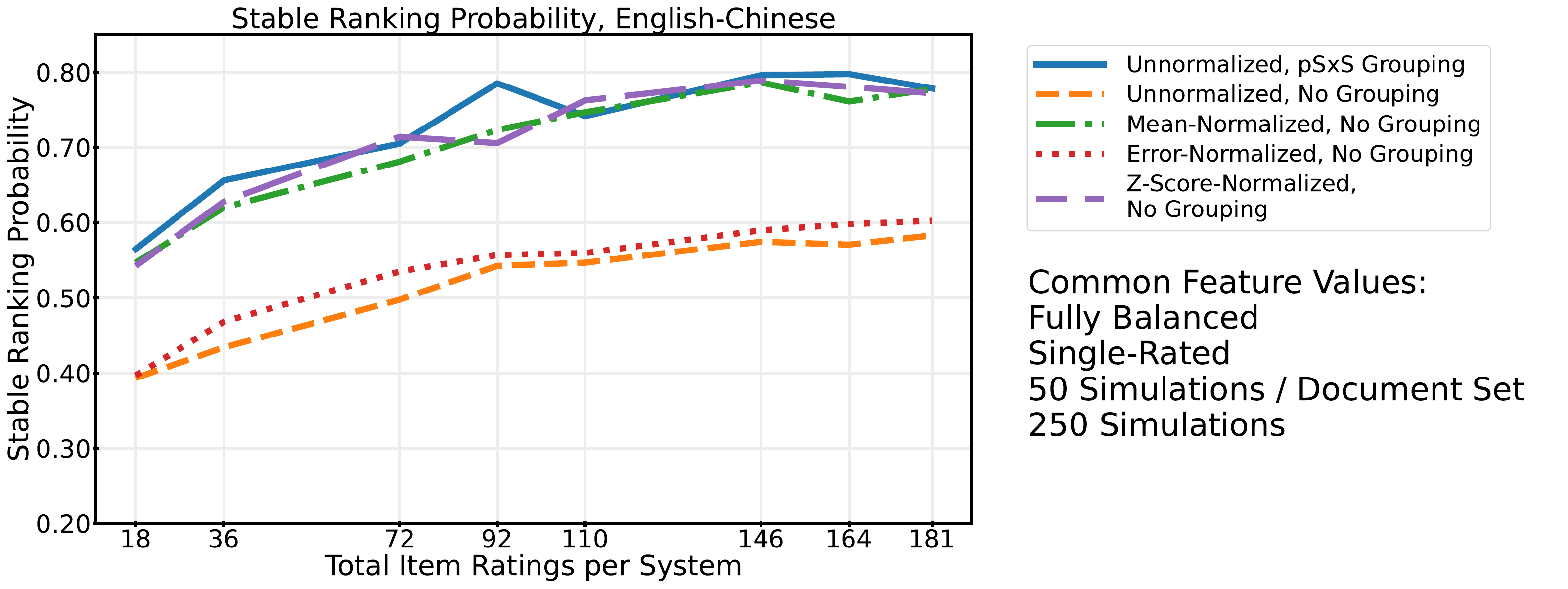}}
    \caption{Evaluation of the Normalization feature in pSxS-grouped vs. ungrouped studies.}
    \label{fig:norm_inval}
\end{figure*}

Figure~\ref{fig:norm_inval} examines the effect of normalization on studies that do not use item grouping, with pSxS-grouped studies shown for reference. Again, Z-score normalization provides a significant stability boost. Mean normalization provides a slightly smaller boost, with error normalization providing little benefit. In English-Chinese, normalization can close the gap below pSxS-grouped studies. 

These latter two experiments show that normalization is especially important when our other recommendations are not followed.

\subsection{\rqrpi}\label{sec:results_numratings}
\begin{figure*}[htb]
    \centering
    \subfloat{\includegraphics[scale=0.18]{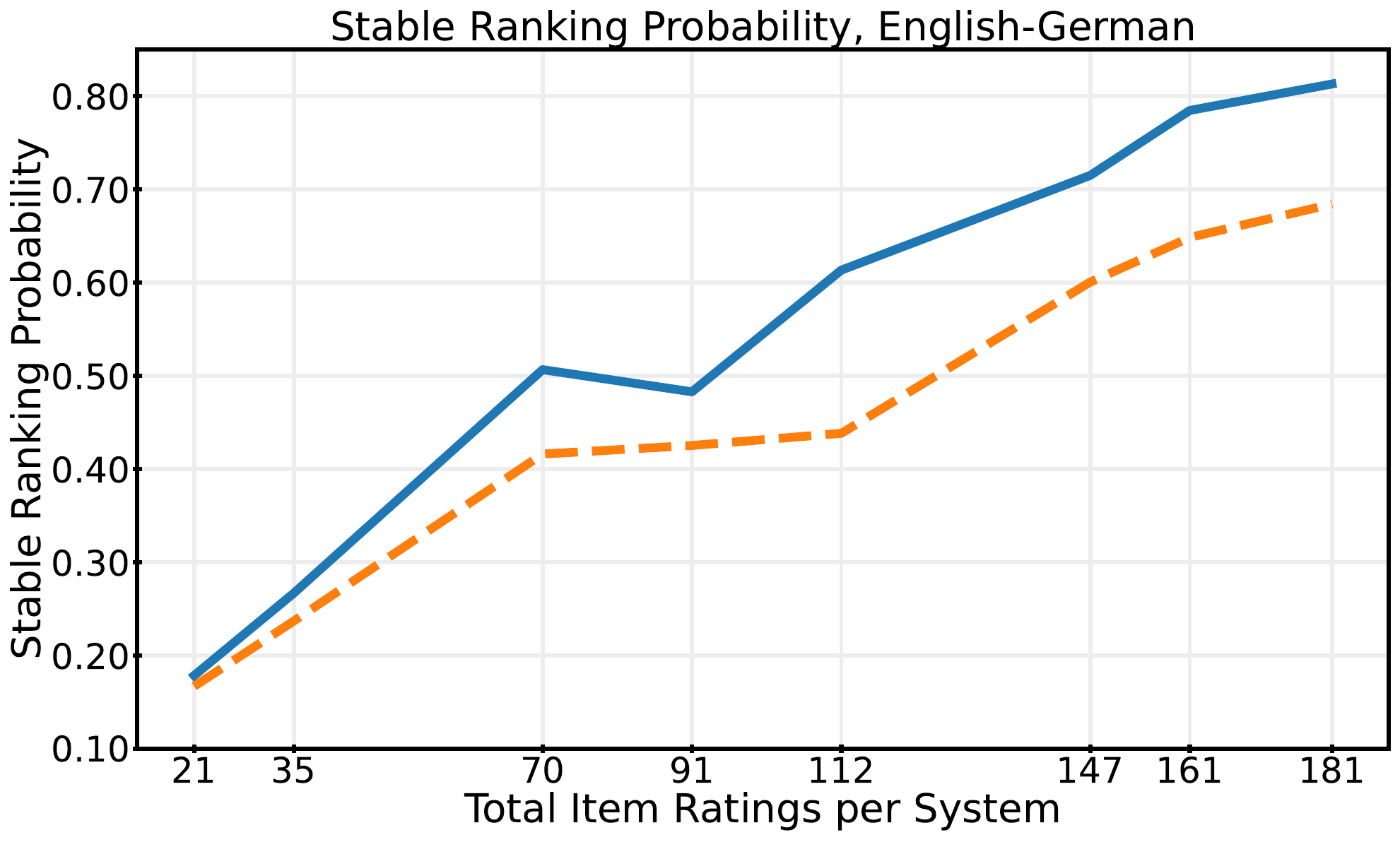}}\hfill
    \subfloat{\includegraphics[scale=0.18]{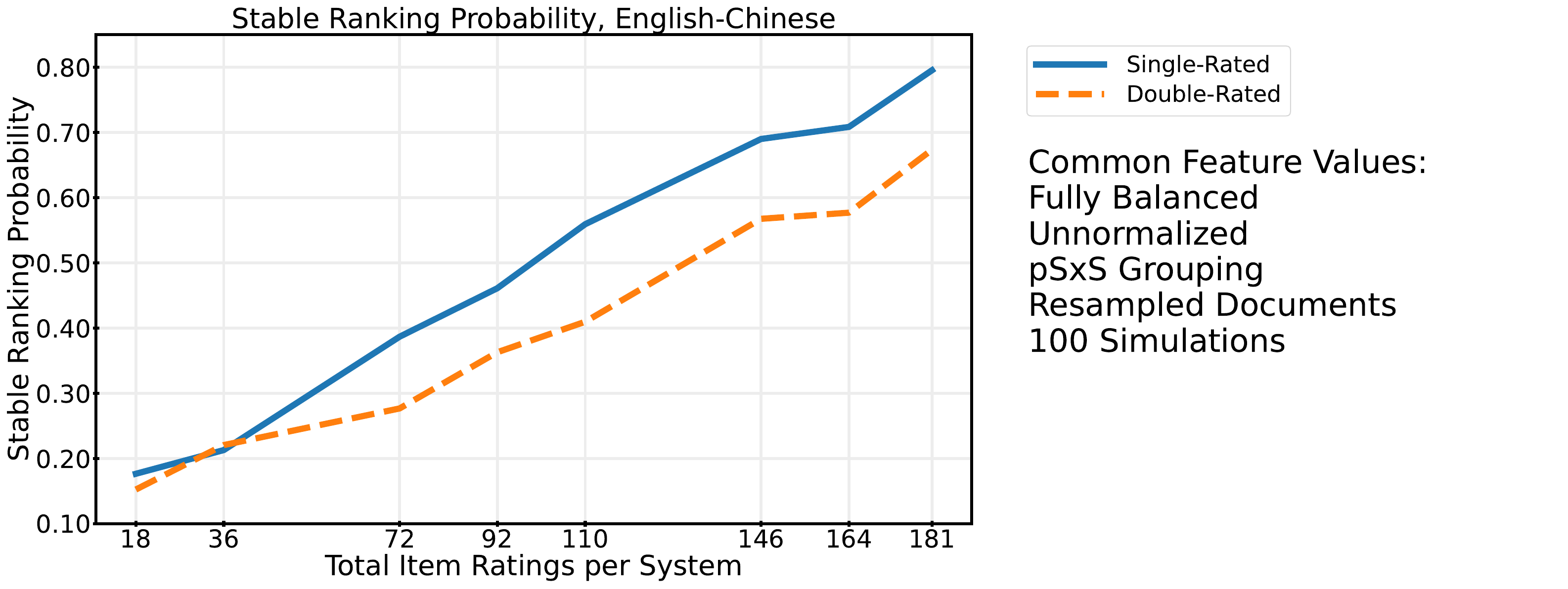}}
    \caption{Evaluation of the Ratings per Item feature, where half as many documents are used in the double-rated case and resampled documents are used in both cases.}
    \label{fig:num_ratings}
\end{figure*}

We make a few modifications to the experimental settings for the analysis of the Ratings per Item feature presented in Figure~\ref{fig:num_ratings}.

First, because we consider a fixed annotation budget, each data point on a double-rated line corresponds to a set of studies each with half as many documents as the single-rated data point at the same position along the $x$-axis. 

Second, we modify the Document Resampling feature to use resampled documents and calculate \favmetricabbr~over all study pairs, not just those with the same set of documents. This is because, when all documents are shared, a pair of simulated double-rated studies can be expected to share twice as many exact ratings as a pair of single-rated studies. This is especially important given that we only have 3 ratings to draw from for each item. By resampling documents, the doubled chance of sharing a rating on a shared document is exactly offset by the halved chance of sharing the document in the first place. Therefore, the expected number of shared ratings between study pairs due to chance is the same in the single- and double-rated case. This results in a fair comparison that essentially measures whether the noise from differences in rater behavior is greater than the noise from differences in relative translation quality for different documents.

Third, we calculate each data point from 100 simulated studies instead of 250. This is because relaxing the restriction of comparing studies with the same documents increases the total number of comparisons, which is approximately offset by the reduction in the number of simulations.

With those considerations covered, we can now conclude from Figure~\ref{fig:num_ratings} that single-rated studies on $n$ documents are consistently more stable than double-rated studies on $\frac{n}{2}$ documents, indicating that annotation budget is best spent on increasing the number of items rather than ratings per item.
\section{Related Work}\label{sec:relwork}

The work of \citet{saldias-fuentes-etal-2022-toward} is broadly similar to ours in that it aims to improve MQM.
However, they focus on predicting individual system scores rather than rankings, and investigate ways to reduce the number of items annotated by a single rater while minimizing error relative to scores derived from the full test set.  \citet{gladkoff2022measuring} construct a statistical model of an MQM-like annotation process, and run simulations with it to estimate confidence intervals for various sample sizes.
In contrast, we characterize the effects of different rater/item configurations and sample sizes, and focus on stability rather than accuracy compared to an assumed ground truth provided by a larger sample. Complementary to our work, \citet{popovic-2021-agree} presents a fine-grained study on how rater agreement varies with error type, and \citet{popovic-belz-2022-reporting} investigate how agreement depends on different methods for normalizing error counts, finding that raw error counts (which we use) provide the most reliable estimates.

Other work has investigated properties of direct Likert-style human scoring for MT with non-expert crowd annotators. Building on earlier proposals for score collection and normalization \cite{graham2013continuous,graham-etal-2014-machine}, \citet{graham-etal-2015-accurate} measure the number of ratings per item required to achieve good correlation with ground-truth scores, concluding that at least 15 ratings are required.
More recently, \citet{knowles-2021-stability} examined the effect on ranking stability of assigning raters to different systems and documents, making recommendations that are consistent with our pSxS proposal.
\citet{wei2022searching} characterize how the number of samples required for adequately-powered statistical tests depends on the magnitude of the difference between system scores,
and suggest strategies to reduce this number in practice. 
Finally, \citet{licht2022consistent} propose a new adequacy-focused rating scheme with better inter-annotator agreement than generic ratings, and a method for calibrating ratings so they are comparable across different languages.

Beyond MT, human assessment of automatically-generated text is an area of increasing research focus. The surveys by \citet{howcroft2020twenty} and \citet{gehrmann2023repairing} provide an excellent guide to work in this area.

As for MT evaluation datasets, \citet{zouhar2024finetuned} released a dataset of 25,000 MQM ratings in the biomedical domain. The most similar dataset to ours is that of \citet{freitag-etal-2021-experts}, consisting of approximately 100,000 MQM ratings on WMT data. In contrast, our dataset contains approximately 140,000 ratings.

\section{Conclusion}
In this work we have proposed a framework for evaluating reliability of NLG evaluations and used it to provide recommendations to practitioners. Our recommendations are supported by analysis of MT data in two language pairs. Additionally, we release these MQM annotations to the public to allow for follow-up research by the community.

\section*{Acknowledgments}
We thank the MQM raters for their work providing the annotations in our dataset. We also thank Jon Clark for his exceptionally helpful review of an early draft of this work. Finally, we thank the anonymous reviewers for their comments, which improved the quality of the final version of the paper.

\clearpage
\section*{Limitations}\label{sec:limitations}
Throughout this work, we refer to our datasets as ``English-German'' and ``English-Chinese'' because that is the most relevant distinction between them, but we caution readers against concluding that differences in results between the two are due to intrinsic properties of German and Chinese. Other differences in our datasets include the systems, raters, bucketing scheme, and document lengths.

The recommendations we make in this work are supported by our analysis and we offer plausible explanations for our results. However, there are variables not accounted for in this work, such as collecting more than 3 ratings per item, increasing the rater pool size, using different MT systems, evaluating documents from different domains, using language pairs other than the two presented here, and others. We therefore cannot guarantee that the same trends will be seen on different datasets.

We also specifically call out the fact that the ratings in our dataset come from expert raters who are very familiar with the task, so our recommendations may be less applicable to settings with non-expert raters.

We only consider system ranking, meaning our analysis does not consider the potential goal of reliably quantifying the absolute quality of a single system. We also consider stability of that ranking, but it is possible that reducing the variance of the ranking could potentially bias it away from the (unknown) ``true'' ranking.

Finally, we use MT and MQM to draw our conclusions. While we expect our recommendations to be generally applicable to other NLG evaluation settings, it is possible that, for tasks other than MT or evaluation frameworks that are highly dissimilar to MQM, our recommendations may be less applicable.
\bibliography{references}

\appendix
\section{Appendix}

\subsection{Dataset Bucketing}\label{app:buckets}
Tables~\ref{tab:ende_buckets} and \ref{tab:enzh_buckets} illustrate the specific bucket-rater assignments used in our datasets.
\begin{table}[htb]
\centering
\footnotesize
\subfloat[English-German]{
\begin{tabular}{cc}
\toprule
\textbf{Bucket \#} & \textbf{Raters}\\
\midrule
1 & A,B,C\\
2 & B,C,D\\
3 & C,D,E\\
4 & D,E,F\\
5 & E,F,G\\
6 & F,G,A\\
7 & G,A,B\\
\bottomrule
\end{tabular}\label{tab:ende_buckets}
}
\quad
\subfloat[English-Chinese]{
\begin{tabular}{cc}
\toprule
\textbf{Bucket \#} & \textbf{Raters}\\
\midrule
1 & H,I,J\\
2 & K,L,M\\
\bottomrule
\end{tabular}\label{tab:enzh_buckets}
}
\caption{Bucket-rater assignments for each dataset. For English-German, raters were rotated through the buckets. For English-Chinese, buckets were rated by 3 raters with no overlap between buckets. English-German buckets contain 25-26 documents, while English-Chinese buckets contain 90-91 documents.}
\end{table}

\subsection{System Mean MQM Scores}\label{app:mean_scores}
Table~\ref{tab:system_means} shows the mean MQM scores of systems included in our datasets.
\begin{table}[htb]
    \centering
    \footnotesize
    \subfloat{
    \begin{tabular}{cc}
         \toprule
         
         \textbf{English-German System} & \textbf{MQM Score}\\
         \midrule
         Online-W&                0.81 \\
         refB&                    0.98 \\
         MBR-bleu&                1.05 \\
         Online-B&                1.09 \\
         JDExploreAcademy&        1.16 \\
         MBR-bleurt&              1.17 \\
         MBR-comet&               1.18 \\
         Online-A&                1.33 \\
         Online-Y&                1.39 \\
         Online-G&                1.40 \\
         QUARTZ&                  1.44 \\
         Lan-Bridge&              1.51 \\
         OpenNMT&                 1.77 \\
         PROMT&                   1.78 \\
         M2M100&                  2.96 \\
         \bottomrule
    \end{tabular}}
    \quad
    \subfloat{
    \begin{tabular}{cccc}
         \toprule
         \textbf{English-Chinese System} & \textbf{MQM Score} \\
         \midrule
         refB&                1.45 \\
         Lan-Bridge&          1.60 \\
         Online-W&            1.67 \\
         JDExploreAcademy&    1.88 \\
         Online-B&            1.91 \\
         LanguageX&           1.97 \\
         Manifold&            2.01 \\
         HuaweiTSC&           2.04 \\
         AISP-SJTU&           2.10 \\
         Online-A&            2.21 \\
         Online-Y&            2.26 \\
         DLUT&                2.57 \\
         Online-G&            2.65 \\
         \bottomrule
    \end{tabular}}
    \caption{Average segment-level MQM scores for systems in both datasets (lower is better). Note that ``refB'' in both datasets is a human reference translation.}
    \label{tab:system_means}
\end{table}

\subsection{Simulating Studies with Specified Properties}\label{app:simulation}
Here we offer some more detail on how we simulate studies from specified properties. As mentioned in Section~\ref{sec:sim_procedure}, Item Grouping and Load Balancing control most aspects of the simulation method. The procedures we used, based on these two properties, are:

    \textbf{pSxS, Fully Balanced}: For each bucket, shuffle the documents and assign them to the 3 raters of that bucket in a round-robin fashion. Each rater rates all items for each assigned document.
    
    \textbf{pSxS, Entropy-Balanced}: Begin with a random rater assignment for each document. Then, in a random order, re-assign each document to the rater that moves the normalized entropy of the overall rater distribution closest to the target; if there are multiple raters that satisfy this, pick one at random. If the resulting study's normalized entropy is outside the tolerance range, reject it and repeat.
    
    \textbf{No Grouping}: For any load balancing scheme, use the same procedure as in the Valid pSxS case, but operate on \textit{items} instead of \textit{documents}.
    
    \textbf{System-Balanced Grouping, Fully Balanced}: For each bucket and each system, shuffle the raters and documents for that bucket and then assign the outputs of that system to the raters in a round-robin fashion.

Incorporating other properties requires only minor changes to the above procedures:
    
    \textbf{Ratings per Item}: For double-rated studies, assign to \textit{pairs} of raters instead of single raters; this is trivial because, at 3 ratings per item, the number of rater pairs is also 3.
    
    \textbf{Number of Documents}: Sub-sample documents from each bucket equally to achieve the target number before rater assignment.
    
    \textbf{Normalization}: Apply the specified normalization after rater assignment.


\end{document}